\documentclass[journal]{IEEEtran}
\usepackage{xcolor}
\usepackage{cite}
\usepackage{amsmath,amssymb,amsfonts}
\usepackage[ruled,linesnumbered]{algorithm2e}
\DontPrintSemicolon
\usepackage{graphicx}
\usepackage{textcomp}
\usepackage{wrapfig}
\usepackage{fontawesome5}
\usepackage[breaklinks=true,hidelinks]{hyperref}
\usepackage{xurl}
\usepackage{needspace}

\def\BibTeX{{\rm B\kern-.05em{\sc i\kern-.025em b}\kern-.08em
    T\kern-.1667em\lower.7ex\hbox{E}\kern-.125emX}}
\begin{document}
\title{In-Ride Alcohol-Impairment Detection in E-Scooterists with False-Alarm Control}
\author{\mbox{Marco Capuccini\textsuperscript{1,2} and Rahul Rajendra Pai\textsuperscript{1,3}}
\thanks{\raggedright \textsuperscript{1}Voi Technology AB, Stockholm, Sweden (e-mail: marco.capuccini@voiapp.io).}
\thanks{\raggedright \textsuperscript{2}Division of Transport Planning, KTH Royal Institute of Technology, Stockholm, Sweden (e-mail: marcap@kth.se).}
\thanks{\raggedright \textsuperscript{3}Division of Vehicle Safety, Department of Mechanical Engineering, Chalmers University of Technology, Gothenburg, Sweden (e-mail: rahul.pai@chalmers.se).}
\thanks{\raggedright (Corresponding author: Marco Capuccini.)}}

\maketitle

\begin{abstract}
Shared e-scooter services have become a widely adopted urban transport mode. While most users ride responsibly, alcohol intoxication stands out among the factors contributing to severe crashes. Nonetheless, countermeasures remain limited to single-point reaction tests and night bans that suspend the service altogether. This paper proposes a new approach in which onboard sensors evaluate the rider as the trip unfolds, raising an alarm as soon as enough evidence of impairment has accumulated. Specifically, we introduce a detector that operates on inertial and throttle measurements, with a provable bound on the rate of false alarms. Experiments on sensor data from 141 rides, in which 25 participants rode while sober and at two target blood alcohol concentration levels, confirm that the bound holds, whereas baselines and ablations either exceed it or lose detection performance, and in some cases delay the alarm. At a bound of 0.023, the detector identifies 91\% of the rides performed at the higher concentration and 50\% of those at the lower one, with median detection times of 25 and 27 seconds, respectively. We further show that an embedded implementation meets the real-time requirement, making mitigation actions feasible onboard---without requiring data to leave the vehicle. Overall, this work lays the ground for interventions that reach impaired riders as soon as possible, sparing the sober ones the burden of a pre-ride test or the suspension of the service at night, while letting operators budget false alarms against user experience.
\end{abstract}


\section{Introduction}
\label{sec:introduction}
\IEEEPARstart{S}{hared} micromobility, in particular rental e-scooters, has transformed cities at a pace unmatched by other forms of transportation. The scale of this shift is evident in Europe, with 312M trips in 2025 alone~\cite{Fluctuo_2026}. As ridership continues to grow, safety has become a pressing challenge, underscored by crash reports~\cite{Stigson_2021}. Among the contributing factors, alcohol intoxication stands out across studies. In Sweden, 44\% of fatal e-scooter crashes recorded between 2016 and 2024 involved intoxicated riders, against 13\% for bicyclists and 27\% for e-cyclists~\cite{Pai_2026}. Non-fatal injury data follow the same pattern, with intoxication reported in 12\% of injured riders in Washington, D.C. during 2019~\cite{Cicchino_2021}, 32\% in a German emergency department during 2022~\cite{Hartz_2025}, and 47\% in Helsinki during 2021, before nighttime restrictions~\cite{Dibaj_2024}. Despite this evidence, preventive interventions remain limited: the in-app reaction tests currently adopted by operators are easily circumvented and screen the rider only at the start of the trip~\cite{Voi_2020}, while nighttime bans suspend the service for everyone, sober riders included~\cite{Dibaj_2024}.

In this paper, we propose an approach in which onboard sensors monitor rider kinematics continuously throughout the ride. This approach removes the need for a pre-ride reaction test, keeps the service available at night, and enables in-ride intervention which occurs only when impairment is detected. For such a system to be viable, we also provide a false-alarm-rate (FAR) control mechanism with an anytime-valid statistical guarantee, allowing operators to bound the rate of incorrect detections and thus limit unnecessary alarms for sober riders. Anytime validity means that the bound holds at every step of the ride rather than at a single pre-specified decision point; the detector can therefore raise an alarm as soon as sufficient evidence accumulates, without inflating the FAR. This extends our prior work, which established alcohol impairment as detectable from the permutation entropy (PE) of sensor signals, though only from complete rides in post-processing and without FAR control~\cite{https://doi.org/10.48550/arxiv.2609.38276}.

The proposed detector evaluates the ride by extracting the PE of triaxial accelerometer, gyroscope, and throttle lever position measurements, computed over a short lookback window. The resulting seven-dimensional vector is then fed to a machine learning (ML) model that yields an impairment score for the current window. As the ride progresses, evidence is accumulated through a running statistic of the scores, and an alarm is raised as soon as it crosses a threshold calibrated on sober rides. The threshold is obtained through a procedure introduced in this work, which builds on inductive conformal prediction (ICP)~\cite{DBLP:conf/ecml/PapadopoulosPVG02}, extending its FAR control to every step of the ride.

We evaluate the detector on the dataset we publicly released with our prior work, comprising signals from 141 instrumented e-scooter rides, in which 25 participants rode while sober and at two target blood alcohol concentration (BAC) levels~\cite{https://doi.org/10.48550/arxiv.2609.38276}. The detector keeps the FAR within the bound, unlike baselines that do not account for the temporal dependence of sensor measurements. At a FAR bound of 0.023, it detects 91\% of the rides at the higher BAC level within a median of 25\,s of riding, and 50\% of the rides at the lower level within a median of 27\,s, a rate no in-ride baseline matches under FAR control. We also run an embedded implementation of the detector on an STM Nucleo board~\cite{ST_2026}, where one detection cycle takes 262\,ms, showing that in-ride monitoring is feasible onboard.

\needspace{16\baselineskip}
\begin{samepage}
In summary, the key contributions of this work are as follows:
\begin{itemize}
    \item We introduce an alcohol impairment detector that monitors e-scooter riding continuously and raises an alarm as soon as sufficient evidence accumulates, reducing the need for pre-ride screening and nighttime bans.
    \item We propose a threshold calibration procedure that statistically bounds the FAR at any step of the ride, and we prove its validity.
    \item Using instrumented e-scooter signals, we empirically show that (1) the detector keeps the FAR within the bound where baselines that disregard temporal dependence do not, (2) it outperforms in-ride baselines under FAR control with strong detection performance at the higher BAC level and moderate performance at the lower, and (3) it runs in real time on an embedded device.
    \item We publicly release the full experimental code.
\end{itemize}
\end{samepage}

\section{Related Work}
\label{sec:related-work}
Alcohol impairment has been studied most extensively in the automotive domain. A systematic review of 26 studies applying ML to in-vehicle sensor data, most often tracking vehicle dynamics and control inputs, reports consistently strong discriminative performance~\cite{Devcich_2026}. In micromobility, impaired riding has been examined for both cyclists and e-scooterists, with the evidence pointing to riding fitness degrading measurably with BAC~\cite{Hartung_2014,Zube_2022}, while detection from onboard sensors has so far been demonstrated only in our prior work~\cite{https://doi.org/10.48550/arxiv.2609.38276}. To our knowledge, threshold calibration procedures and anytime-valid FAR control, enabling in-trip detection, remain unexplored across both domains.

ICP provides a computationally efficient mechanism to provably bound the FAR in ML systems, without prior distributional knowledge~\cite{DBLP:conf/ecml/PapadopoulosPVG02}. This property is particularly compelling, as in real applications the underlying distribution is seldom known and guarantees derived from approximations can be unreliable. ICP validity instead holds under exchangeability alone~\cite{Barber_2023}, weaker than the i.i.d. sampling which ML methods commonly assume. Exchangeability martingales, first developed to test this condition~\cite{DBLP:conf/icml/VovkNG03}, provide valid FAR control and have also been applied to anomaly detection in flight behavior~\cite{DBLP:conf/mdm/HoSSSK19} and, in our prior work, to e-scooter GNSS traces~\cite{DBLP:conf/copa/CapucciniPC25}. However, when readings are temporally dependent, as in sensor streams, exchangeability breaks even in the absence of anomalies and the FAR guarantee no longer holds. The flight behavior study neither discusses this nor reports the empirical FAR, whereas our prior work addresses it with a shuffling buffer, which restores exchangeability at the cost of a considerable detection delay. Conformal methods that instead relax the condition also fall short for in-ride detection, as weighted quantiles bound the loss of coverage rather than removing it~\cite{Barber_2023}, and block permutation is only approximately valid under dependence~\cite{DBLP:conf/colt/ChernozhukovWZ18}, both targeting coverage at a single test point rather than anytime validity.

\section{Detection Method}
\label{sec:method}
The detector operates on rides consisting of sensor readings. Formally, a ride $R=\{\mathbf{x}(t_k)\}_{k=1}^{T}$ is a discrete-time signal of length $T$ sampled at a rate $F_s=100$\,Hz, with
\begin{equation*}
    \mathbf{x}(t_k)\in\mathbb{R}^d,\quad k=1,\dots,T,
\end{equation*}
where $d = 7$ is the number of sensor channels, i.e., triaxial acceleration, triaxial angular velocity, and throttle position. The inertial readings pass through a causal 64th-order finite impulse response low-pass filter, using the Hamming window at a 10\,Hz cutoff~\cite{Oppenheim_2010}, which suppresses chassis vibration while preserving voluntary motor input~\cite{Winter_2009}. As the ride is monitored while it unfolds, the detector can in principle access all the readings up to the current step $t_k$. However, for resource efficiency, it retains only a short lookback window of $w=1000$ samples, updated every $\Delta=100$ samples, i.e., once per second. It follows that the resulting window stream is given by
\begin{equation*}
    W_j=\{\mathbf{x}(t_k)\}_{k=k_j-w+1}^{k_j},\quad k_j=w+(j-1)\Delta,
\end{equation*}
for $j=1,\dots,J$ and $J=\lfloor(T-w)/\Delta\rfloor+1$.

Features are extracted from each window as the normalized PE of its channels~\cite{Bandt_2002}, with embedding dimension 5 and a delay matched to the Nyquist rate of the filtered signal. Letting $H$ denote PE and $W_j^{(1)},\dots,W_j^{(d)}$ the channels of $W_j$, the stream of feature vectors is defined as
\begin{equation*}
    \mathbf{z}_j=\big(H(W_j^{(1)}),\dots,H(W_j^{(d)})\big)\in\mathbb{R}^d,\quad j=1,\dots,J.
\end{equation*}
To account for subject variability, as in our prior work~\cite{https://doi.org/10.48550/arxiv.2609.38276}, each $\mathbf{z}_j$ is then centered by subtracting a channel-wise mean, giving
\begin{equation*}
    \mathbf{c}_j=\mathbf{z}_j-\bar{\mathbf{z}},\quad \bar{\mathbf{z}}=\big(\bar{z}^{(1)},\dots,\bar{z}^{(d)}\big),
\end{equation*}
where $\bar{z}^{(1)},\dots,\bar{z}^{(d)}$ are the channel means of the feature vectors over a set of centering rides from the same rider. Note that this operation requires no knowledge of their impairment level, as it only removes a rider-specific offset.

The remainder of this section describes inference and threshold calibration, then proves the anytime-valid FAR guarantee.

\subsection{Inference}
\label{sec:inference}

\begin{algorithm}[t]
\caption{In-Ride Inference}
\label{alg:inference}
\KwIn{Window stream $W_1, W_2, \dots, W_J$; scoring function $f$; running evidence function $g$; subject center $\bar{\mathbf{z}}$; threshold $\tau_\alpha$}
\For{$j \in \{1, 2, \dots, J\}$}{
    $\mathbf{c}_j \gets \big(H(W_j^{(1)}),\dots,H(W_j^{(d)})\big) - \bar{\mathbf{z}}$\label{lin:features}\;
    $s_j \gets f(\mathbf{c}_j)$\label{lin:score}\;
    $E_j \gets g(s_1,\dots,s_j)$\label{lin:statistic}\;
    \If{$E_j > \tau_\alpha$\label{lin:test}}{
        Raise an alarm~\faBell{} and exit\label{lin:alarm}\;
    }\label{lin:endtest}
}
\end{algorithm}

Algorithm~\ref{alg:inference} shows how inference is carried out on the stream. The procedure requires a scoring function $f$, which rates each window such that higher values indicate stronger evidence of impairment. For this, we fit a logistic regression on the centered feature vectors of a set of training rides and take its impairment probability as the score, a modeling choice that in our prior work matched more complex alternatives at a lower computational cost~\cite{https://doi.org/10.48550/arxiv.2609.38276}. In contrast with the centering rides, training rides may come from other riders, though their impairment level must be known. Inference further requires a running evidence function $g$, chosen as the shrunk average
\begin{equation*}
    g(s_1,\dots,s_j)=\frac{1}{j+\kappa}\Big(\sum_{i=1}^{j}s_i+\kappa s_0\Big),
\end{equation*}
where $s_1,\dots,s_j$ are the scores of the windows observed so far and the shrinkage $\kappa=10$ acts as a number of pseudo-windows pulling the average toward the sober baseline $s_0$, i.e., the mean score over sober rides used for fitting, thereby damping small-sample fluctuations early in the ride. The remaining inputs are the subject center $\bar{\mathbf{z}}$ and the threshold $\tau_\alpha$, which results from the calibration in Section~\ref{sec:calibration} for a FAR bound $\alpha$.

The procedure runs one cycle per window, as these become available, with $J$ only known when the ride ends. First, it extracts and centers the features (line~\ref{lin:features}), then scores them (line~\ref{lin:score}) and folds the score into the running evidence $E_j$ (line~\ref{lin:statistic}). If $E_j$ exceeds $\tau_\alpha$ (line~\ref{lin:test}), an alarm is raised and the procedure exits (line~\ref{lin:alarm}), otherwise it proceeds to the next cycle. As the shrunk average can be computed from a running sum, resource utilization does not grow with $j$.

\subsection{Threshold Calibration}
\label{sec:calibration}
Let $R_1,\dots,R_K$ be sober calibration rides sampled at random, disjoint from the training and centering rides. Each $R_i$, $i=1,\dots,K$, is processed as in Algorithm~\ref{alg:inference}, without testing for impairment (lines~\ref{lin:test}--\ref{lin:endtest} disabled), and is summarized by the peak of its running evidence
\begin{equation*}
    S_i=\max_{j=1,\dots,J_i}g\big(f(\mathbf{c}_1),\dots,f(\mathbf{c}_j)\big),
\end{equation*}
where $\mathbf{c}_1,\dots,\mathbf{c}_{J_i}$ are the centered feature vectors of $R_i$, with $J_i$ denoting its total number of windows. Writing the peaks in increasing order as $S_{(1)}\le\dots\le S_{(K)}$, the threshold is set to
\begin{equation*}
    \tau_\alpha=S_{(\lceil(K+1)(1-\alpha)\rceil)},
\end{equation*}
namely the empirical $(1-\alpha)$ quantile of the calibration peaks, where the count includes the ride yet to be observed and the rank is rounded up. In the next section we prove that testing against $\tau_\alpha$, as in lines~\ref{lin:test}--\ref{lin:endtest} of Algorithm~\ref{alg:inference}, keeps the FAR within $\alpha$ at every step of the ride.

\subsection{Anytime-Valid False-Alarm Guarantee}
\label{sec:guarantee}
The argument in this section borrows the validity proof of ICP~\cite{DBLP:conf/ecml/PapadopoulosPVG02}, reproduced in brief to keep this work self-contained. Our contribution is the reduction of a ride to the peak of its running evidence, which lifts the bound from a single test point to every step of the ride.

Let $S=\max_j E_j$, $j=1,\dots,J$, denote the peak of the ride under test, exchangeable with the calibration rides and therefore sober, i.e., any $E_j$ crossing the threshold yields a false alarm. As $S\ge E_j$ at every step, such a crossing occurs if and only if $S>\tau_\alpha$, hence
\begin{equation*}
    \mathbb{P}\big(\exists\, j:\, E_j>\tau_\alpha\big)=\mathbb{P}\big(S>\tau_\alpha\big).
\end{equation*}
Since $f$, $g$, and the centering are fixed before the calibration rides are drawn, the peak is a deterministic function of a ride, so $S_1,\dots,S_K,S$ inherit the exchangeability of the rides they summarize, ensured by random sampling for $R_1,\dots,R_K$ and assumed for the ride under test. Ties among the peaks almost surely do not occur for a continuous score, and would in any case only make the test more conservative~\cite{DBLP:conf/nips/TibshiraniBCR19}. Setting them aside, $S$ is therefore equally likely to occupy any of the $K+1$ ranks, giving
\begin{equation}
\label{eq:rank}
    \mathbb{P}\big(S>\tau_\alpha\big)=\frac{K+1-\lceil(K+1)(1-\alpha)\rceil}{K+1}\le\alpha.
\end{equation}
The anytime-valid bound follows directly:
\begin{equation*}
    \mathbb{P}\big(\exists\, j:\, E_j>\tau_\alpha\big)=\mathbb{P}\big(S>\tau_\alpha\big)\le\alpha.
\end{equation*}

\section{Detection Performance}
\label{sec:performance}
In the previous section we have shown the FAR to be bounded by construction, without saying anything about how often impairment is detected, nor how quickly, both of which depend on the scoring function $f$ and on the running evidence function $g$. This section confirms the bound empirically and evaluates detection rate and time, in either case against baselines with and without FAR control guarantees. The dataset, the metrics, the experimental setup, and the baselines are described next, followed by the results.

\subsection{Dataset}
\label{sec:dataset}
The evaluation relies on the dataset released with our prior work~\cite{https://doi.org/10.48550/arxiv.2609.38276}, gathered in a controlled experiment approved by the Swedish Ethical Review Authority (ref.\ 2025-08327-02). Thirty-three participants rode a commercial e-scooter limited to 7\,km/h on an indoor track combining a straight line, a slalom with narrowing cone spacing, a braking zone, and a figure-eight. Before data collection, they practiced until they felt comfortable with the vehicle and the layout. Each then covered the track twice while sober, followed by two trials at each of two target BAC levels in increasing order, 0.05\% and 0.08\%, confirmed with a law-enforcement-grade breathalyzer before every trial. The vehicle logged the channels listed in Section~\ref{sec:method} at 100\,Hz. Intermittent memory exhaustion of the onboard logger led to the exclusion of eight participants and the loss of nine trials across eight others, leaving 25 (19 male), aged $26.5\pm4.6$ years. The resulting dataset contains 141 rides: 45 sober, 50 at the lower level, and 46 at the higher one; at least one trial per participant is retained in each condition. Missing samples remain within these recordings, fewer than 10\% in any of them, which we fill by linear interpolation. We refer the reader to~\cite{https://doi.org/10.48550/arxiv.2609.38276} for the full protocol.

\subsection{Evaluation Metrics}
\label{sec:metrics}
We measure performance in terms of FAR, detection rate, and detection time, at different values of the bound $\alpha$. Given a set of test rides $\mathcal{R}=\{R_1,\dots,R_N\}$, the first two are computed as the alarm rate
\begin{equation*}
    \frac{\big|\{\,R\in\mathcal{R}:\, E_j>\tau_\alpha\ \text{for some}\ j\,\}\big|}{N},
\end{equation*}
where $E_j$, $j=1,\dots,J$, is the running evidence of $R$. Specifically, $\mathcal{R}$ contains sober rides for the FAR and impaired ones for the detection rate. Detection time is instead computed, on each impaired ride raising an alarm, as
\begin{equation*}
    \frac{w+(j^\ast-1)\Delta}{F_s},\quad j^\ast=\min\{\,j:\, E_j>\tau_\alpha\,\},
\end{equation*}
which amounts to the ride time elapsed when the alarming window $j^\ast$ closes.

\subsection{Experimental Setup}
\label{sec:setup}
To make the most of the available data, the evaluation follows a leave-one-participant-out protocol. At every fold, a distinct scoring function is fit on the rides of the participants left in, whose sober rides also provide the baseline $s_0$, whereas centering is carried out within each participant, one ride at a time against the remaining ones. For the participant left out, a distinct threshold is calibrated on the sober rides left in, each contributing the peak it reaches under the scoring function of the fold that holds it out, so that no calibration peak is in-sample. Each ride is therefore scored once, out of fold, giving one running evidence path per ride and one threshold per participant, over which the metrics above are computed. Furthermore, for both rates we derive a 95\% confidence interval (CI) from 1000 bootstrap resamples of the out-of-fold results, drawn such that the rides of each participant stay together, whereas detection time is summarized by its median and interquartile range (IQR). Note that no leakage occurs, as no ride is ever scored by a function fit on data from the participant who rode it, whose other rides enter only through the label-free centering.

In this protocol, the scoring function and the threshold vary from one fold to the next, hence each threshold is computed from peaks produced by distinct models. Each scoring function nonetheless comes from the same deterministic procedure, applied to all participants but the one left out. It follows that the peaks do not depend on the order in which the participants are taken, thereby inheriting their exchangeability. Rides are however not drawn at random, as those of the participant under test are left out. We argue that centering, by removing rider-specific offsets, makes this distinction negligible. The argument in Section~\ref{sec:guarantee} thus applies within each fold, and the bound carries over to the pooled metrics, as Section~\ref{sec:results} confirms. The same reasoning covers the CIs, as resampling draws participants at random.

Finally, we point out that the smallest attainable $\alpha$ follows from the size of the calibration set. Holding out a participant leaves at least $K=43$ sober rides, placing the tightest threshold at the $1/(K+1)\approx0.023$ level. Hence, we show results for $\alpha$ set to 0.023, 0.05, 0.1, and 0.2 in Section~\ref{sec:results}, higher values being of no practical interest.

\subsection{Baselines}
\label{sec:baselines}
We compare the detector against five baselines under the protocol introduced in the previous section. First, we ablate in-ride detection and the calibrated running evidence, to quantify what each contributes to the evaluation metrics. Then, we turn to exchangeability martingales, to assess whether they control FAR in practice under time dependence, and how they compare once recalibrated. Finally, we ablate and sweep the shrinkage parameter. We leave the window size at $w=1000$ samples, which strikes a good balance between PE estimation accuracy and the minimum detection delay. Each baseline is detailed next.

\subsubsection{Batch Classifier}
This baseline ablates in-ride detection, thus reproducing our prior work~\cite{https://doi.org/10.48550/arxiv.2609.38276}. In summary, the scoring function is fit on PE computed over the whole ride without windowing, and the ride is flagged as impaired at test time when its score exceeds $1-\alpha$, with no calibration involved. Detection time is undefined, as the alarm cannot precede the end of the ride.

\subsubsection{Naive Windowed Classifier}
We re-introduce in-ride detection, and thus windowing, but keep running evidence and calibration ablated. Specifically, the scoring function is fit on windowed PE and tests each window separately, with the ride flagged as impaired as soon as the score exceeds $1-\alpha$.

\subsubsection{Exchangeability Martingale}
The window scores in the proposed detector are turned into smoothed conformal p-values against the sober calibration rides, and the Simple Jumper strategy bets on them sequentially as in~\cite{DBLP:conf/copa/VovkPNACG21}, with jump probability 0.01. The resulting wealth process is tested against $1/\alpha$, following Ville's inequality~\cite{Ville_1939}.

\subsubsection{Recalibrated Martingale}
The wealth process introduced above serves as running evidence in the proposed detector, with the threshold calibrated as in Section~\ref{sec:calibration} instead of taken from Ville's inequality.

\subsubsection{Shrinkage Sweep}
The proposed detector is run with the shrinkage $\kappa$ set to 0, 10, 20, and 30, the first of which ablates the pull toward the sober baseline.

\subsection{Results}
\label{sec:results}

\begin{table*}[t]
\caption{False Alarm Rate at Four Bounds, With 95\% CIs in Brackets}
\label{tab:far}
\centering
\renewcommand{\arraystretch}{1.3}
\begin{tabular*}{0.82\textwidth}{@{\extracolsep{\fill}}lcccc}
\hline
Method & $\alpha=0.023$ & $\alpha=0.05$ & $\alpha=0.1$ & $\alpha=0.2$\\
\hline
Proposed detector & 0.02 [0.00--0.07] & 0.04 [0.00--0.11] & 0.09 [0.02--0.17] & 0.18 [0.07--0.29]\\
Batch classifier & 0.02 [0.00--0.07] & 0.02 [0.00--0.07] & 0.02 [0.00--0.07] & 0.02 [0.00--0.07]\\
Naive windowed classifier & 0.11 [0.04--0.20] & 0.27 [0.13--0.42] & 0.44 [0.30--0.58] & 0.80 [0.66--0.93]\\
Exchangeability martingale & 0.67 [0.52--0.80] & 0.71 [0.58--0.84] & 0.78 [0.64--0.91] & 0.87 [0.77--0.95]\\
Recalibrated martingale & 0.02 [0.00--0.07] & 0.04 [0.00--0.11] & 0.11 [0.02--0.22] & 0.18 [0.07--0.31]\\
\hline
\end{tabular*}
\end{table*}

\begin{table*}[t]
\caption{Detection Rate and Median Time at $\alpha=0.023$, With 95\% CIs in Brackets and IQRs in Parentheses}
\label{tab:detection}
\centering
\renewcommand{\arraystretch}{1.3}
\begin{tabular*}{0.82\textwidth}{@{\extracolsep{\fill}}lcccc}
\hline
& \multicolumn{2}{c}{0.08\% BAC} & \multicolumn{2}{c}{0.05\% BAC}\\
Method & Rate & Time, s & Rate & Time, s\\
\hline
Proposed detector ($\kappa=10$) & 0.91 [0.82--0.98] & 25 (22--29) & 0.50 [0.36--0.64] & 27 (22--35)\\
Batch classifier & 0.96 [0.88--1.00] & --- & 0.24 [0.12--0.36] & ---\\
Recalibrated martingale & 0.65 [0.50--0.79] & 54 (52--57) & 0.14 [0.04--0.24] & 60 (56--64)\\
\hline
\multicolumn{5}{l}{Shrinkage sweep of the proposed detector}\\
\hline
$\kappa=0$ & 0.63 [0.48--0.78] & 10 (10--10) & 0.22 [0.12--0.34] & 10 (10--10)\\
$\kappa=20$ & 0.89 [0.80--0.96] & 29 (26--33) & 0.46 [0.32--0.60] & 34 (28--42)\\
$\kappa=30$ & 0.85 [0.74--0.94] & 33 (30--37) & 0.38 [0.24--0.52] & 41 (33--50)\\
\hline
\end{tabular*}
\end{table*}

Table~\ref{tab:far} shows the FAR attained at the four levels of $\alpha$. The proposed detector stays within the bound throughout, with point estimates consistently below $\alpha$, and never significantly above it---as the lower CI ends show. Since the guarantee holds in probability and not in realization, the upper ends do lie above $\alpha$, two to three times at the two lowest bounds, though expected of an estimate over only 45 sober rides. The recalibrated martingale holds the bound as well, save at $\alpha=0.1$, where one sober ride beyond what the bound allows brings the point estimate to 0.11, an excess its lower CI end leaves insignificant. This is unsurprising, as this baseline inherits the full procedure introduced in this work, but for the running evidence, keeping the anytime-valid guarantee intact. It is also worth noting that for both variants the point estimates fall close to the bound, as one would expect from \eqref{eq:rank}, where the rate falls short of $\alpha$ only by the rounding up of the rank. Conversely, the batch classifier, while keeping the FAR well below $\alpha$, at 0.02 with the lower CI end at 0.00 throughout, yields overly conservative predictions, which cost detection rate at lower BAC level, as shown later at $\alpha=0.023$. Finally, both the naive windowed classifier and the exchangeability martingale significantly inflate the FAR, with point estimates several times the bound, and lower CI ends above it. This confirms that testing each window on its own, or betting on time-dependent p-values, forfeits FAR control.

Table~\ref{tab:detection} shows detection rate at $\alpha=0.023$, for the methods that control the FAR. The proposed detector alarms on 91\% of the rides at the higher BAC level, within a median of 25\,s of riding (IQR 22--29\,s), and on 50\% of those at the lower level, within a median of 27\,s (IQR 22--35\,s). Conversely, the recalibrated martingale falls short, detecting only 65\% and 14\% of them, respectively, at medians of 54\,s and 60\,s of riding, with its CIs never overlapping those of the proposed detector. Lastly, the batch classifier detects 96\% of the rides at the higher BAC level, a margin that the strong overlap of their CIs suggests is insignificant. At the lower level, its conservative predictions bring detection down to 24\%, its CI meeting our detector's at 0.36 without overlapping.

We conclude this section by discussing the shrinkage sweep shown in the bottom half of Table~\ref{tab:detection}. When shrinkage is turned off, i.e., $\kappa=0$, most alarms arrive at the earliest step the window affords, 10\,s at either BAC level. This comes at the cost of detection rate, decreasing to 63\% and 22\%, with neither CI overlapping those of the proposed detector; the fluctuations triggering these early alarms also inflate the sober peaks that set the threshold. For the other values considered, the rate peaks at $\kappa=10$, the value chosen in this work. It then recedes as the pull toward the sober baseline grows, down to 85\% and 38\% at $\kappa=30$, where the median alarm is further delayed at both levels. The FAR stays at 0.02 across the sweep.

\section{On-Device Latency}
\label{sec:on-device}
To assess the feasibility of real-time detection, we implement Algorithm~\ref{alg:inference} in C, and evaluate it bare metal on a NUCLEO-H533RE board~\cite{ST_2026}, at the default 32\,MHz system clock. Specifically, we time 1000 cycles against the 1\,ms system tick. In each, the incoming samples are filtered and accumulated into a window of size $w=1000$, with the procedure then carrying on as described, up to the update of the running evidence, where the cycle ends. As specific values do not bear on the size of the computation, both the learned parameters and the input signal are drawn at random within their valid ranges, outside of the timed region.

Over the 1000 cycles, latency averages 262.26\,ms, with a standard deviation of 0.91\,ms and a range from 260 to 265\,ms. A cycle thus takes about a quarter of the second that separates consecutive windows, leaving generous margin for the real-time requirement.

\section{Discussion and Limitations}
\label{sec:discussion}
This paper proposes a new approach in which riders are evaluated for impairment as they ride, rather than at a single point, before or after the trip. Evidence builds from riding itself, reducing the need for pre-ride screening, yet the alarm comes as soon as enough of it has accumulated. The rate at which sober rides raise one is itself a design parameter, with a provably valid guarantee, allowing operators to budget the cost in riding experience beforehand. The calibration procedure assumes only knowledge of sober outcomes, matching the reality of production-grade implementations, where an adequate number of impaired rides cannot be collected due to scarce reporting and ethical concerns. In practice, we envision the scoring function fit as a one-class model on rides from periods in which impairment is unlikely, weekday mornings for instance, with the threshold calibrated on a large number of disjoint rides from the same periods---lowering the smallest attainable bound well below 0.023. This would likely cause a small degree of contamination, though only inflating peaks at the cost of detection rate, leaving the guarantee effectively intact. Centering is equally undemanding, as it needs only a handful of prior rides from the rider, none of which carry labels.

Ethical considerations bear on how an alarm should be acted upon. Interventions should be tiered, transparent, and subject to human review where the response carries a significant cost for the rider. In fact, a single alarm signals an atypical riding pattern with respect to the sober baseline, not a determination of intoxication. It should thus inform soft measures, e.g., a warning or a speed reduction, rather than accusation or sanction. Such measures can be taken onboard, as Section~\ref{sec:on-device} shows, with no data leaving the vehicle, thus mitigating privacy concerns.

While we presented encouraging results, some limitations should be acknowledged. First, detection performance appears to be stronger at the higher BAC level, that is 0.08\%. We argue that, as the aim is to reduce the number of serious injuries, such a concentration should still be considered low: among fatally injured riders with a positive BAC in Sweden, the median ranged from 0.12\% to 0.20\% across modes~\cite{Pai_2026}. However, detection at such levels remains untested, as inducing that degree of intoxication would expose participants to a higher risk of injury. At the same time, ground truth requires controlled dosing under ethical approval, which puts naturalistic labels out of reach and keeps datasets such as the one used in this study small. The leave-one-participant-out protocol makes the most of what is available, yet cannot push the bound below 0.023 or narrow the CIs (see Section~\ref{sec:results}). The data-collection setup also leaves us with a fixed repertoire of maneuvers on a single vehicle, whereas traffic, slopes, and surfaces add variation in real-world scenarios, which may raise the sober peaks or mask impairment. Recalibrating per vehicle and geographical location is however inexpensive, hence the FAR can effectively be kept under control; precise predictions would then follow from running the detector where and when intoxicated riding is suspected to be prevalent.

\section{Conclusion}
\label{sec:conclusion}
This paper presents an alcohol impairment detector that monitors e-scooterists as riding unfolds, raising an alarm as soon as enough evidence accumulates. The FAR is turned into a design parameter and bounded at every step of the ride, a guarantee we prove and confirm empirically, whereas baselines that disregard the temporal dependence of sensor measurements forfeit it. Detection is strong at the higher BAC level and moderate at the lower one, with alarms typically raised within half a minute of riding. In addition, we show that an embedded implementation leaves ample margin for real-time operation, allowing data to stay on the vehicle. The public release of the experimental code supports reproducibility and further work on the topic. Overall, the proposed detector provides a foundation for interventions that reach impaired riders while the trip is still underway, reducing the need for pre-ride screening and nighttime bans.

\section*{Data and Code Availability}
All data and experimental code required to reproduce the reported results are publicly available at \url{https://github.com/voi-oss/in-ride-impairment-detection} and \url{https://doi.org/10.6084/m9.figshare.34070439}.

\section*{Acknowledgments}
We thank Rahman Amandius from Voi Technology for his valuable feedback on this paper. We also thank Sven \r{A}kersten and Mia Muminovic from Voi Technology for their help with the embedded implementation.

The data used in this paper come from the experiment reported in our prior work~\cite{https://doi.org/10.48550/arxiv.2609.38276}, run at MicroLab, Chalmers University of Technology, Sweden. We sincerely thank the participants for their time, Marco Dozza, Niyathi Kini, Alexander Rasch, and Ali Mohammadi from Chalmers for the data collection, and Andrea de Bejczy at the University of Gothenburg for medical support and oversight. We are also grateful to Rikard Karlsson and Tord Hansson for access to the Eventhallen, as well as to Henrik Hörlin, Peter Bäckgren, and Kristina Henricson Briggs for the Tracks facilities.

The data collection experiment was supported by the MicroTox project, funded by VINNOVA (Sweden’s innovation agency) and Drive Sweden under grant 2025-00431, the MinTOX project, funded by the Area of Advance Transport at Chalmers, and Stöd till MinTOX, funded by Trafikverket under grant TRV2024/106354. This work was partially supported by the Wallenberg AI, Autonomous Systems and Software Program (WASP) funded by the Knut and Alice Wallenberg Foundation.

\section*{Disclosure of Interests}
The authors have no competing interests to declare that are relevant to the content of this article.

\section*{Use of Generative AI}
Generative AI tools were used to assist with information retrieval, experimental code development, and text editing. The authors take full responsibility for the content of this paper.

\bibliographystyle{IEEEtran}
\bibliography{refs}

\end{document}